\documentclass{article} 
\usepackage{iclr2021_conference,times}

\usepackage{amsmath,amsfonts,bm}

\def\eqref#1{equation~\ref{#1}}

\def\1{\bm{1}}

\def\eps{{\epsilon}}

\def\rvx{{\mathbf{x}}}

\def\vg{{\bm{g}}}

\def\vi{{\bm{i}}}

\def\vx{{\bm{x}}}

\DeclareMathAlphabet{\mathsfit}{\encodingdefault}{\sfdefault}{m}{sl}
\SetMathAlphabet{\mathsfit}{bold}{\encodingdefault}{\sfdefault}{bx}{n}

\def\gL{{\mathcal{L}}}

\newcommand{\R}{\mathbb{R}}

\DeclareMathOperator*{\argmax}{arg\,max}
\DeclareMathOperator*{\argmin}{arg\,min}

\usepackage{hyperref}
\usepackage{url}

\usepackage{times}
\usepackage{epsfig}
\usepackage{graphicx}
\usepackage{array} 
\usepackage{booktabs} 
\usepackage{makecell} 
\usepackage{multirow} 
\usepackage{caption}
\usepackage{color}
\usepackage{xcolor}
\usepackage{paralist}
\usepackage{comment}
\usepackage{amsthm,amsmath,amssymb}
\usepackage{subfigure}
\usepackage{wrapfig,lipsum,booktabs}
\usepackage{enumitem}
\usepackage{color}
\usepackage{cases}
\usepackage{algorithm2e}
\RestyleAlgo{ruled}
\newtheorem{proposition}{Proposition}
\newtheorem{definition}{Definition}
\usepackage{enumitem}

\usepackage{color}

\title{How and When Adversarial Robustness Transfers in Knowledge Distillation?}

\author{Rulin Shao\\
Carnegie Mellon University\\
\texttt{rulins@cs.cmu.edu} \\
\And
Jinfeng Yi\\
JD AI Research\\
\texttt{yijinfeng@jd.com}
\And
Pin-Yu Chen\\
IBM Research\\
\texttt{pin-yu.chen@ibm.com}
\And
Cho-Jui Hsieh\\
University of California, Los Angeles\\
\texttt{chohsieh@cs.ucla.edu}
}

\iclrfinalcopy 
\begin{document}

\maketitle

\begin{abstract}
    Knowledge distillation (KD) has been widely used in teacher-student training, with applications to model compression in resource-constrained deep learning. Current works mainly focus on preserving the accuracy of the teacher model. However, other important model properties, such as adversarial robustness, can be lost during distillation. This paper studies how and when the adversarial robustness can be transferred from a teacher model to a student model in KD. 
    We show that standard KD training fails to preserve adversarial robustness, and we propose KD with input gradient alignment (KDIGA) for remedy.
    Under certain assumptions, 
    we prove that the student model using our proposed KDIGA can achieve at least the same certified robustness as the teacher model.
    Our experiments of KD contain a diverse set of teacher and student models with varying network architectures and sizes evaluated on ImageNet and CIFAR-10 datasets, including residual neural networks (ResNets) and vision transformers (ViTs). Our comprehensive analysis shows several novel insights that (1) With KDIGA, students can preserve or even exceed the adversarial robustness of the teacher model, even when their models have fundamentally different architectures; (2) KDIGA enables robustness to transfer to pre-trained students, such as KD from an adversarially trained ResNet to a pre-trained ViT, without loss of clean accuracy; and (3) Our derived local linearity bounds for characterizing adversarial robustness in KD are consistent with the empirical results.
\end{abstract}

\section{Introduction}

Knowledge distillation (KD)~\citep{hinton2015distilling, distillation1} is a popular machine learning framework for teacher-student training, with appealing applications to model compression in resource-constrained deep learning~\citep{sun2019patient, wang2019private}, such as memory-efficient inference on edge or mobile devices~\citep{wang2021knowledge, lyu2020differentially}. In essence, KD trains a small model under the supervision of a large teacher model with the goal of improving or retaining the performance of the student model. For classification tasks, existing works mainly focus on preserving the accuracy of the teacher model~\citep{zagoruyko2016paying, passban2020alp, mirzadeh2020improved}, while ignoring other important properties, such as adversarial robustness.
For a student model, failing to preserve the same level of adversarial robustness as the teacher model can bring about a false sense of successful knowledge distillation when put into deployment. Therefore, ensuring and improving the adversarial robustness of the student model is critical to the safe deployment of the model in many practical applications. 

\begin{figure}[t]
    \centering
    \includegraphics[width=0.9\columnwidth]{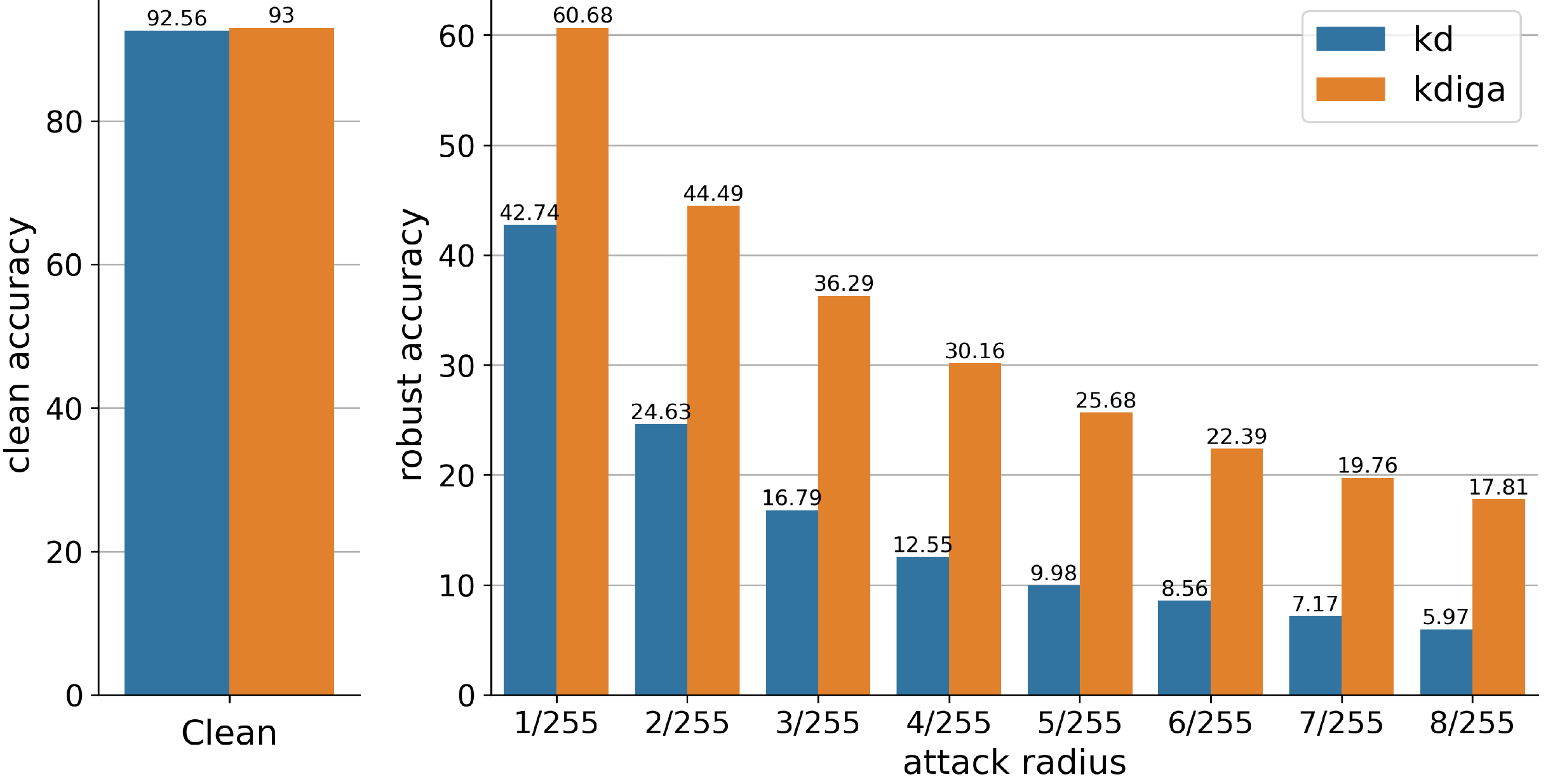}
    \caption{Clean accuracy (\%) and robust accuracy (\%) of the student models (MobileNetV2) against 20-step PGD attack \citep{pgd-attack} with different radii on CIFAR-10. The two students are distilled from the same adversarially trained WideResNet with TRADES \citep{trades}. ``KD'' stands for the standard knowledge distillation and ``KDIGA'' stands for our proposed knowledge distillation with input gradient alignment. 
    }
    \label{fig:barplot}
\end{figure}

To illustrate the critical but overlooked failure mode of standard KD,
in Figure\ref{fig:barplot} we show that it cannot preserve the adversarial robustness of the teacher model, and propose to use input gradient alignment in KD (we name it KDIGA) for better adversarial robustness preservation. In addition to empirical evidence, in this paper we also prove that our method can make the student achieve at least the same certified robustness as the teacher model under certain assumptions. When comparing our method with other baselines on ImageNet~\citep{imagenet} and CIFAR-10~\citep{cifar10} datasets, the results show substantial improvement in the adversarial robustness of the student models obtained by our method. 

To demonstrate the generality of our proposed KDIGA method, we further study the transferability of adversarial robustness between convolutional neural networks (CNNs)~\citep{resnet} and vision transformers (ViTs)~\citep{vit}. We show that our method enables the transfer of adversarial robustness between these two fundamentally different architectures. We also show that KDIGA can improve the adversarial robustness of a pre-trained ViT without sacrificing the clean accuracy.
We also extend our theoretical analysis and use local linearity measures to characterize the transfer of adversarial robustness in KD, and show that our derived performance bounds match the trends of the empirical robustness.
\paragraph{Our Contributions}
\begin{itemize}[leftmargin=*]
    \item We propose to use KD with input gradient alignment (KDIGA) to train both accurate and adversarially robust student models in knowledge distillation. For instance, using KDIGA, the robust accuracy of the student model can be significantly increased from 5.97\% to 17.81\% compared with KD on CIFAR-10, with even better clean accuracy, as shown in Figure~\ref{fig:barplot}. On ImageNet, the robust accuracy of the student model is increased from 1.5\% to 37.5\% using KDIGA compared with KD.
    \item We show that adversarial robustness can be transferred between fundamentally different architectures with KDIGA, i.e., the robust accuracy of ResNet18 distilled from ViTs can achieve or even exceed the teacher's robust accuracy.
    \item KDIGA also extends to pre-trained student models. When we distill from adversarially trained ResNet50 to the normally pre-trained ViT in a fine-tuning approach, the robust accuracy of ViT boosts up to $11.1\times$ larger, together with even higher clean accuracy on the ImageNet dataset. We also find that students with higher learning capacity can achieve better results.
    \item We prove that the student model distilled with KDIGA can achieve at least the same certified robustness as the teacher with some mild assumptions. We further generalize the analysis and provide a bound with local linearity measures for characterizing adversarial robustness in KD, which is consistent with the empirical results on ImageNet and CIFAR-10.
\end{itemize}

\section{Related Work}
There are some recent works studying when and how adversarial robustness will transfer in different machine learning settings, such as transfer learning~\citep{hendrycks2019using, chen2020adversarial, shafahi2019adversarially}, representation learning~\citep{chan2020thinks} and Model-agnostic meta-learning (MAML)~\citep{wang2021on}. In contrast,  we focus on the setting of knowledge distillation.
The basic Knowledge Distillation (KD) formulates the supervised learning objective as
\begin{equation}
    \argmin_{f^s} \mathcal{L}_{KD}(\vx, y) = \argmin_{f^s} \lambda_{CE}\mathcal{L}_{CE}(f^s(\vx), y)+\lambda_{KD}T^2\mathcal{L}_{KL}(f^s(\vx)/T,f^t(\vx)/T)
\end{equation}
where $f^s$ is the student model , $f^t$ is the teacher model, $(\vx, y)\in\mathcal{D}$, $\mathcal{D}$ is the training set, $\mathcal{L}_{CE}$ is the cross-entropy loss, $\mathcal{L}_{KL}$ is the KL-divergence loss, $\lambda_{CE}$ and $\lambda_{KD}$ are constant factors to balance the two losses, and $T$ is a temperature factor. 

One effective way to train adversarially robust model is adversarial training~\citep{pgd-attack,trades,imagenet-robsut-cnn}, which adds adversarial perturbations to the inputs during training and forces the model to learn robust predictions. \cite{ard} follows the same idea and formulates an adversarially robust distillation (ARD) objective using adversarial training:
\begin{equation}\label{eq:ard}
    \argmin_{f^s} \mathcal{L}_{ARD}(\vx, y) = \argmin_{f^s} \lambda_{CE}\mathcal{L}_{CE}(f^s(\vx), y)+\lambda_{KD}T^2\mathcal{L}_{KL}(f^s(\vx+\delta)/T,f^t(\vx)/T),
\end{equation}
\begin{equation}\label{equ:delta}
    \text{and}~~\delta = \argmax_{\|\delta\|_p \leq \epsilon}\mathcal{L}_{CE}(f^s(\vx+\delta), y).
\end{equation}
However, it is computationally expensive to calculate the adversarial perturbations for all training data especially when the dataset is large-scale (e.g. the ImageNet dataset~\citep{imagenet}). There are some major differences between our method and ARD. Firstly, our method does not use adversarial training and thus is much more computationally efficient. Secondly,  Our method can also be used together with ARD and further improve the robust accuracy of the student model.

Projected gradient descent (PGD) is one of the most commonly used adversarial attacks for both adversarial robustness evaluation and adversarial training, which solves Eq.~\ref{equ:delta} by iteratively taking gradient ascent by
\begin{equation}
\rvx_{t+1}^{a d v}=Clip_{\rvx_0,\eps}(\rvx_{t}^{a d v}+\alpha \cdot \operatorname{sgn}\left(\nabla_{\rvx} \mathcal{L}_{CE} \left(\rvx_{t}^{a d v}, y\right)\right)),
\end{equation}
where $t=1,\cdots, T$, $T$ is the number of iterations, $\rvx^{adv}_t$ stands for the solution after $t$ iterations, $\nabla_x$ denotes the gradient with respect to $x$, and $Clip_{\rvx_0,\eps}(\cdot)$ denotes clipping the values to make each $ \rvx_{t+1}^{a d v}$ within $ [\rvx_{0}-\eps,\rvx_{0}+\eps]$, according to the $\ell_p$ norm bounded threat model. The adversarial perturbation is then obtained by $\delta_{\text{pgd}} = \rvx_T^{a d v} - \rvx_0$. In addition, AutoAttack (\cite{auto-attack}) is currently the strongest white-box attack which evaluates adversarial robustness with a parameter-free ensemble of diverse attacks.
\section{Knowledge Distillation with Input Gradient Alignment}

In this section, we first introduce our proposed framework of knowledge distillation with input gradient alignment (KDIGA) which we find is critical for adversarial robustness preservation in knowledge distillation. Then we prove that the student model can achieve at least the same certified robustness as the teacher model under two assumptions. We also give a general bound to analyze the factors that affect the transferability of adversarial robustness in KDIGA.

\subsection{Problem Formulation}\label{sec:formulation}

Suppose $f^s(\vx):\R^D \rightarrow \R^N$ is the student model and $f^t(\vx):\R^D \rightarrow \R^N$ is the teacher model, where $D$ is the input dimension and $N$ is the number of classes. In KDIGA, we force the student to learn both the logits and gradient knowledge from the teacher model, so the objective is defined as:
\begin{equation}
\begin{aligned}
    \argmin_{f^s} \mathcal{L}_{IGA}(\vx, y) =&\argmin_{f^s} \left[ \lambda_{CE}\mathcal{L}_{CE}(f^s(\vx), y)+\lambda_{KL}T^2\mathcal{L}_{KL}(f^s(\vx)/T,f^t(\vx)/T)\right.\\
    &\left.+\lambda_{IGA}\|\nabla_{\vx}\mathcal{L}_{CE}(f^s(\vx),y)-\nabla_{\vx}\mathcal{L}_{CE}(f^t(\vx),y\|_2)\right], 
\end{aligned}
\end{equation}
where $(\vx, y)\in\mathcal{D}$ is the input image and the corresponding label in the training dataset, $\mathcal{L}_{CE}$ and $\mathcal{L}_{KL}$ stand for the cross-entropy loss and the KL-divergence loss respectively, $\lambda_{CE}, \lambda_{KL}$ and $\lambda_{IGA}$ are constants that balance the trade-off between different losses, $T$ is the temperature factor, and $\|\cdot\|_2$ is the $\ell_2$ norm. The pseudo code of KDIGA is shown in Algorithm~\ref{alg:kdiga}.

\begin{algorithm}[t]
\caption{Pseudocode of KDIGA}\label{alg:kdiga}
\KwIn{teacher $f^t$, student $f^s_{\theta}$ with trainable parameters $\theta$, training set $\mathcal{D}$,$\lambda_{CE}$,  $\lambda_{KL}$, $\lambda_{IGA}$, learning rate $\eta$, number of epochs $N_{epochs}$.}
\KwOut{adversarially robust student $f^s_{\theta}$.} 
\For{$epoch \in N_{epochs}$}{
\For{batch $(\vx, y) \in \mathcal{D}$}{
    $p_s, p_t \gets f_{\theta}^s(\vx), f^t(\vx)$\;
    $\ell_s, \ell_t \gets \mathcal{L}_{CE}(p_s, y), \mathcal{L}_{CE}(p_t, y)$\;
    $\ell_{KL} \gets T^2 \mathcal{L}_{KL}(p_s/T, p_t/T)$\;
    $g_s, g_t \gets \nabla_{\vx}\ell_s, \nabla_{\vx}\ell_t$\;
    $\ell_{iga} \gets \lambda_{CE}\ell_s + \lambda_{KL}\ell_{KL} + \lambda_{IGA}\|g_s-g_t\|_2$\;
    $\theta \gets \theta - \eta\nabla_{\theta}\ell_{iga}$\;
}
}
\end{algorithm}

Besides, we show two ways to combine our method with adversarial training strategies for KD using ARD \citep{ard}, i.e., KDIGA-ARD$_C$ and KDIGA-ARD$_A$. The objective for them are
\begin{equation}
\begin{aligned}
    \argmin_{f^s} \mathcal{L}_{IGA_C}(\vx, y) =&\argmin_{f^s} \left[ \lambda_{CE}\mathcal{L}_{CE}(f^s(\vx), y)+\lambda_{KL}T^2\mathcal{L}_{KL}(f^s(\vx+\delta)/T,f^t(\vx)/T)\right.\\
    &\left.+\lambda_{IGA}\|\nabla_{\vx}\mathcal{L}_{CE}(f^s(\vx),y)-\nabla_{\vx}\mathcal{L}_{CE}(f^t(\vx),y\|_2)\right],
\end{aligned}
\end{equation}
\begin{equation}
\begin{aligned}
    \argmin_{f^s} \mathcal{L}_{IGA_A}(\vx, y) =&\argmin_{f^s} \left[ \lambda_{CE}\mathcal{L}_{CE}(f^s(\vx), y)+\lambda_{KL}T^2\mathcal{L}_{KL}(f^s(\vx+\delta)/T,f^t(\vx+\delta)/T)\right.\\
    &\left.+\lambda_{IGA}\|\nabla_{\vx}\mathcal{L}_{CE}(f^s(\vx+\delta),y)-\nabla_{\vx}\mathcal{L}_{CE}(f^t(\vx+\delta),y\|_2)\right],
\end{aligned}
\end{equation}
where ``$IGA_C$'' is in short for KDIGA-ARD$_C$ and ``$IGA_A$'' is in short for KDIGA-ARD$_A$, $\delta$ is calculated by Eq.~\ref{equ:delta} as inner maximization. KDIGA-ARD$_C$ is a direct combination of the original ARD formulation in Eq.~\ref{eq:ard} with our proposed IGA loss on clean samples as an additional regularization. KDIGA-ARD$_A$ further considers perturbed samples in IGA. Their key difference is that KDIGA-ARD$_C$ only aligns predictions of student on perturbed samples with the predictions of teacher on clean samples in the KL-divergence loss.
On the other hand, KDIGA-ARD$_A$ forces the student to align with both the predictions and input gradients of the teacher on perturbed samples. We also tried other variants but did not observe notable differences.

\subsection{Preservation of Certified Robustness}

In this section, we prove that using KDIGA, the student model can provably achieve as good  robustness as the the teacher model's in ideal situations. We first formally define $\delta$-robust in Definition~\ref{def:delta-robust}. 

\begin{definition}\label{def:delta-robust}
($\delta$-robust) Classifier $f(\vx):\R^D \rightarrow \R^N$ is $\delta$-robust if
\begin{equation}
    \argmax f(\vx + \bm{\epsilon}) = \argmax f(\vx), \ \ \forall x \in \mathcal{D}, \forall \epsilon \in [\bm{0}, \bm{\delta}]^D.
\end{equation}
\end{definition}

Under mild assumptions, we aim to show that if the teacher model has a  robust radius of $\delta$, then the student model is at least $\delta$-robust under ideal situations. The first assumption is the {\it perfect student} assumption in which we 

suppose $f^s:\R^D \rightarrow \R^N$ is a student model distilled from the teacher model $f^t:\R^D \rightarrow \R^N$ using distillation loss $\mathcal{L}$, and $f^s$ is a perfect student if 
\begin{equation}
    \mathcal{L}(\vx, y) = 0, \forall (\vx, y) \in \mathcal{D},
\end{equation}
which means the student model trust the teacher and can perfectly learn the knowledge defined by the distillation objective. The second is the {\it local linearity} assumption, which assumes that neural networks with piece-wise linear activation functions are locally linear (\cite{linearity-local, linearity-max, linearity-towards}) and the certified robust area falls into these piece-wise linear regions.
These two assumptions collaboratively build an ideal situation of knowledge distillation in which we can derive a strong property of KDIGA that the certified robustness of the student model can be as good as or even better than that of the teacher model. Proposition~\ref{pro:robustness} concludes our statement.

\begin{proposition}\label{pro:robustness}
Suppose the teacher model $f^t:\R^D \rightarrow \R^N$ is $\delta$-robust, $f^s:\R^D \rightarrow \R^N$ is a perfect student trained using KDIGA, then $f^s$ is at least $\delta$-robust.
\end{proposition}

We give a proof for Proposition~\ref{pro:robustness} and illustrate why the knowledge distillation without input gradient alignment cannot preserve the adversarial robustness in Appendix~\ref{app:proof1}.

\subsection{General Bound for the Adversarial Robustness of the Student Model}
In this section, we derive a general bound for the adversarial robustness of the student model in knowledge distillation. No assumption is needed for this bound, and the knowledge distillation method is not limited to any specific one. To derive this bound, we first introduce the Local Linearity Measure (LLM, \cite{llm}) in Definition~\ref{def:llm}.

\begin{definition}\label{def:llm}
(Local Linearity Measure) The local linearity of a classifier $f(\vx):\R^{D} \rightarrow \R^{N}$ is measured by the maximum absolute difference between the cross-entropy loss and its first-order Taylor expansion in the $\delta$-neighborhood:
\begin{equation}
    \text{LLM}(f, \vx, \delta) = \max_{\bm{\epsilon} \in B(\delta)} \left| \mathcal{L}_{CE}(f(\vx+\bm{\epsilon}) - \mathcal{L}_{CE}(f(\vx)) - \bm{\epsilon}^T \nabla_{\vx} \mathcal{L}_{CE}(f(\vx)) \right|.
\end{equation}
\end{definition}

\begin{proposition}\label{pro:bound}
Consider a student model $f^s:\R^{D} \rightarrow \R^{N}$ distilled from a teacher model $f^t:\R^{D} \rightarrow \R^{N}$, then $\forall \bm{\epsilon}\in B(\delta)$,
\begin{equation}
    \left| \mathcal{L}_{CE} (f^s(\vx+\bm{\epsilon}),y) - \mathcal{L}_{CE}(f^t(\vx+\bm{\epsilon}),y) \right| \leq \gamma^s  + \gamma^t + \phi
\end{equation}
where $\gamma^s=\text{LLM}(f^s,\vx,\delta)$, $\gamma^t=\text{LLM}(f^t,\vx,\delta)$, and $\phi =  \mathcal{L}_{CE}(f^s(\vx),y) +\mathcal{L}_{CE}(f^t(\vx),y) + \delta \|\nabla_{\vx}\mathcal{L}_{CE}(f^s(\vx),y)-\nabla_{\vx}\mathcal{L}_{CE}(f^t(\vx),y)\|$, and $\| \cdot \|$ is a norm.
\end{proposition}

The proof for Proposition~\ref{pro:bound} can be found in Appendix~\ref{app:proof2}.

Proposition~\ref{pro:bound} states that the adversarial robustness of the student model can be bounded by the LLM of both the student model and the teacher model, the clean accuracy of the student model, and the alignment of the student input gradient with the teacher input gradient. We will use this to further analyze the performance of different knowledge distillation methods in Section~\ref{sec:exp}.
\section{Experiments}\label{sec:exp}
In this section, the ImageNet \citep{imagenet} and CIFAR-10~\citep{cifar10} datasets are used for model training and performance evaluation.

\subsection{Settings}\label{sec:setting}

\vspace{-0.1cm}
\paragraph{Teacher Models} We use pre-trained and publicly available neural networks of varying architectures as teacher models.
For the ImageNet dataset, we use both adversarially trained CNNs and normally trained vision transformers (ViTs) as the teacher models. We use the checkpoint of ResNet50~\citep{resnet} provided by \citet{imagenet-robsut-cnn} which is adversarially trained with an attack radius of $4/255$. We also incorporate ViTs~\citep{vit} as teacher models because they are shown to have better adversarial robustness than CNNs~\citep{robust_vit1,robust_vit2,robust_vit3}, and we are interested in the transferability of adversarial robustness between different architectures. For the CIFAR-10 dataset, we use the WideResNet~\citep{wideresnet} adversarially trained with TRADES~\citep{trades} as the teacher model.

\vspace{-0.1cm}
\paragraph{Student Models} For the ImageNet dataset, we mainly use ResNet18~\citep{resnet} as the student model for experiments . To study the effect of model size, we also consider ResNet34, ResNet50 and ResNet101.
In addition, we use ViT-S/16~\citep{vit} as the student model to study the transferability of adversarial robustness from a CNN teacher to a ViT student.
Unless specified, the student models are all trained from scratch.   Because the training of ViT relies on the large-scale pre-training~\citep{vit}, we use the pre-trained version provided by~\citet{timm} and apply the knowledge distillation methods as a fine-tuning process. For the  CIFAR-10 dataset, we use MobileNetV2~\citep{mobilenetv2} as the student model.

\vspace{-0.1cm}
\paragraph{Training Configurations}
For knowledge distillation on the ImageNet dataset, we run all distillation for $50$ epochs with a batch size of $128$, an initial learning rate of $0.1$ for training from scratch and $0.00001$ for fine-tuning, with milestones at $[20, 30, 40]$ of a decreasing rate of $0.1$. The SGD optimizer with $0.9$ momentum is used to update the model parameters, and a weight decay of $0.0001$ is applied. For basic knowledge distillation, we set the temperature to $1$ and the coefficients of the cross-entropy loss and KL-divergence loss both to $0.5$. For KDIGA, we keep the same setting as that of the basic KD, and set the coefficient of the input gradient alignment term to $\frac{10^3}{B}$, where $B$ is the batch size of the inputs.
For experiments on the CIFAR-10 dataset, we run distillation for $200$ epochs with a batch size of $125$, an initial learning rate of $0.1$ with milestones at $[100, 150]$ of a decreasing rate of $0.1$. The SGD optimizer with a momentum of $0.9$ and a weight decay of $0.0002$ is used to update the parameters. We set the coefficients of the cross-entropy loss and the KL-divergence loss both to $0.5$, and the coefficient for the input gradient alignment to $\frac{10}{B}$, where $B$ is the batch size.

\vspace{-0.1cm}
\paragraph{Evaluation Metrics}
Using the test sets of ImageNet and CIFAR-10, we report the best standard accuracy and the robust accuracy against adversarial attacks of the student models. We conduct $\ell_\infty$ norm bounded adversarial perturbations to generate adversarial examples for evaluating robust accuracy (the pixel value is scaled between 0 to 1), where we use a 40-step projected gradient descent (PGD) attack~\citep{pgd-attack} and the parameter-free AutoAttack~\citep{auto-attack} for 1000 ImageNet test samples, and a 20-step PGD attack and the AutoAttack for all CIFAR-10 test samples. Different attack radii are used to test the robustness of the model under different degrees of adversarial perturbations.

\vspace{-0.1cm}
\paragraph{Notation of Comparative Methods} We denote the standard knowledge distillation method as ``KD'', knowledge distillation combined with adversarial training proposed by \citet{ard} as ``ARD'', our method as ``KDIGA'', and the two kinds of combinations of KDIGA and ARD defined in Section~\ref{sec:formulation} as ``KDIGA-ARD$_C$'' and ``KDIGA-ARD$_A$''. ``Teacher $\xrightarrow{\text{Method}}$ Student'' stands for the distillation from the ``Teacher'' to the ``Student'' using ``Method''.

\subsection{Results on the ImageNet Dataset}~\label{sec:imagenet}
We compare our method with the standard training (ST) and knowledge distillation (KD) on the ImageNet dataset. We find that adversarially robust distillation (ARD) proposed by~\cite{ard} cannot generalize to large-scale dataset, which shows no convergence in the setting as described in Section~\ref{sec:exp} with a very low training speed. So we only compare with ARD in the experiments on CIFAR-10 in Section~\ref{sec:cifar}. We run all experiments on ImageNet for 50 epochs to save training cost and expect better performance can be achieved with more training epochs, e.g. 100 epochs.

Table~\ref{tab:pgd-imagnet} shows the robust accuracy of the models trained using different training strategies, i.e., standard training (ST), standard knowledge distillation (KD) and knowledge distillation with input gradient alignment (KDIGA). Table~\ref{tab:auto-imagnet} supplements some results of KDIGA against AutoAttack, as it is the strongest attack method in the current literature. We show the trend in the AutoAttack result is similar to that of PGD. From these results, we conclude the following observations.

\begin{table}[t]
    \centering
    \caption{Robust accuracy ($\%$) of student models against 40-step PGD attack with different radii and clean accuracy ($\%$) on the ImageNet dataset. Robust accuracy of the teacher models are shown in brackets. The pre-trained student model is denoted with ``*'' where the distillation is conducted as a fine-tuning process. Other students are all trained from scratch. ``ST'' means the model is trained following the standard approach without distillation nor adversarial training. ``AT'' means the model is obtained by adversarial training.}
        \label{tab:pgd-imagnet}
          \vspace{-2mm}
    \resizebox{\columnwidth}{!}{
    \begin{tabular}{l c c c c r}\toprule[1pt]
     & &\multicolumn{4}{c}{\bf{PGD Attack radius}}\\
    \bf{Model} &\bf{Clean} &\bf{0.001} &\bf{0.003} &\bf{0.005} &\bf{0.01}\\ \midrule
    ResNet18 (ST)& 68.7 (-)& 24.9 (-)& 2.0 (-) & 0.6 (-) & 0.0 (-) \\
    ViT-S/16  (ST)& 77.6 (-) & 55.4 (-)& 24.6 (-)& 10.2 (-)& 1.0 (-)\\
    ViT-S/16 (ST)$\xrightarrow{\text{KD}}$ResNet18& 69.0 (77.6)& 30.1 (55.4)& 2.9 (24.6)& 0.6 (10.2)& 0.0 (1.0)\\
    ViT-S/16 (ST)$\xrightarrow{\text{KDIGA}}$ResNet18& 60.0 (77.6)& 51.0 (55.4)& 32.7 (24.6)& 18.0 (10.2)& 3.3 (1.0)\\
    ViT-B/16 (ST)$\xrightarrow{\text{KDIGA}}$ResNet18& 64.7 (76.3)& 52.8 (48.9)& 26.6 (14.6)& 11.3 (6.0)& 0.7 (0.9)\\
    ViT-L/16 (ST)$\xrightarrow{\text{KDIGA}}$ResNet18& 65.9 (80.0)& 53.2 (55.1)& 28.6 (23.4)& 12.4 (9.9)& 1.4 (1.8)\\
    DEIT-S/16 (ST)$\xrightarrow{\text{KDIGA}}$ResNet18& 63.6 (77.7)& 53.1 (48.9)& 31.5 (17.6)& 15.6 (7.1)& 1.6 (1.1)\\
    ResNet50 (AT)$\xrightarrow{\text{KD}}$ResNet18& 66.3 (63.1)& 25.7 (61.9)& 1.5 (59.4)& 0.0 (57.2)& 0.0 (49.0)\\
    ResNet50 (AT)$\xrightarrow{\text{KDIGA}}$ResNet18& 54.2 (63.1)& 48.2 (61.9)& 37.5 (59.4)& 26.5 (57.2)& 9.2 (49.0)\\
    ResNet50 (AT)$\xrightarrow{\text{KDIGA}}$ResNet34& 59.2 (63.1)& 53.9 (61.9)& 42.7 (59.4)& 31.2 (57.2)& 12.1 (49.0)\\
    ResNet50 (AT)$\xrightarrow{\text{KDIGA}}$ResNet50& 58.8 (63.1)& 53.7 (61.9)& 42.2 (59.4)& 31.9 (57.2)& 12.4 (49.0)\\
    ResNet50 (AT)$\xrightarrow{\text{KDIGA}}$ResNet101& 60.3 (63.1)& 55.3 (61.9)& 44.7 (59.4)& 33.1 (57.2)& 12.7 (49.0)\\
    ResNet50 (AT)$\xrightarrow{\text{KDIGA}}$ViT-S/16$^*$& 77.7 (63.1)& 65.3 (61.9)& 50.4 (59.4)& 33.5 (57.2)& 11.1 (49.0)\\
    \bottomrule[1pt]
    \end{tabular}
}
\end{table}

\vspace{-0.1cm}
\paragraph{Standard Knowledge Distillation Cannot Preserve Adversarial Robustness}
As shown in Table~\ref{tab:pgd-imagnet}, models trained using standard training are vulnerable to adversarial perturbations. 
The standard knowledge distillation shows no preservation of adversarial robustness from teacher models, e.g., ResNet18 distilled from ViT-S/16 and the adversarially trained ResNet50 still have low robust accuracy against PGD attack with various radii. When the PGD attack radius is $0.003$, the robust accuracy of the student distilled from \{ResNet50 (AT), ViT-S/16\} is only \{$1.5\%, 2.9\%$\} compared with the teacher's \{$59.4\%, 24.6\%$\}, where ``AT'' is in short for adversarial training. For the attack radius of $0.005$, the robust accuracy of the student distilled from \{ResNet50 (AT), ViT-S/16\} is only \{$0.0\%, 0.6\%$\} compared with the teacher's \{$57.2\%, 10.2\%$\}. 

\vspace{-0.1cm}
\paragraph{Input Gradient Alignment Makes Students More Robust}
From Table~\ref{tab:pgd-imagnet} and Table~\ref{tab:auto-imagnet}, students distilled using KDIGA have higher robust accuracy than those distilled with KD or trained with ST. When the teacher model is ResNet50(AT) and the PGD attack radius is $0.003$, ResNet18 distilled with KDIGA has a robust accuracy of $37.5\%$, while the ResNet18 distilled with KD only has a robust accuracy of $1.5$\%, and ResNet18 (ST) only has a robust accuracy of $2.0$\%. When the teacher model is ViT-S/16 (ST) and the PGD attack radius is $0.003$, ResNet18 distilled with KDIGA has a robust accuracy of $32.7\%$, while the ResNet18 distilled with KD only has a robust accuracy of $2.9$\%. This shows that our proposed input gradient alignment plays the key role to help preserve the adversarial robustness during knowledge distillation.

\begin{table}[t]
    \centering
    \caption{Robust accuracy ($\%$) of student models against AutoAttack with different radii and clean accuracy ($\%$) on the ImageNet dataset. Robust accuracy of the teacher models are shown in brackets. The pre-trained student model is denoted with ``*'' where the distillation is conducted as a fine-tuning process. Other students are all trained from scratch. ``ST'' means the model is trained following the standard approach without distillation nor adversarial training. ``AT'' means the model is obtained by adversarial training.}
        \label{tab:auto-imagnet}
          \vspace{-2mm}
    \resizebox{\columnwidth}{!}{
    \begin{tabular}{l c c c c r}\toprule[1pt]
     & &\multicolumn{4}{c}{\bf{AutoAttack Attack radius}}\\
    \bf{Model} &\bf{Clean} &\bf{0.001} &\bf{0.003} &\bf{0.005} &\bf{0.01}\\ \midrule
    ResNet18  (ST)& 68.7 (-) & 14.3 (-)& 0.4 (-)& 0.0 (-)& 0.0 (-)\\
    ViT-S/16  (ST)& 77.6 (-) & 48.1 (-)& 6.0 (-)& 0.5 (-)& 0.0 (-)\\
    ViT-S/16 (ST)$\xrightarrow{\text{KDIGA}}$ResNet18& 60.0 (77.6)& 47.2 (48.1)& 25.0 (6.0)& 10.1 (0.5)& 0.7 (0.0)\\
    ViT-B/16 (ST)$\xrightarrow{\text{KDIGA}}$ResNet18& 64.7 (76.3)& 49.6 (39.8)& 19.4 (5.4)& 5.0 (0.6)& 0.0 (0.0)\\
    ViT-L/16 (ST)$\xrightarrow{\text{KDIGA}}$ResNet18& 65.9 (80.1)& 49.6 (46.6)& 19.1 (8.5)& 5.8 (1.0)& 0.0 (0.0)\\
    DEIT-S/16 (ST)$\xrightarrow{\text{KDIGA}}$ResNet18& 63.6 (80.1)& 50.0 (0.4)& 23.7 (0.0)& 7.8 (0.0)& 0.1 (0.0)\\
    ResNet50 (AT)$\xrightarrow{\text{KDIGA}}$ResNet18& 54.2 (63.1)& 45.9 (47.5)& 31.9 (42.5)& 19.1 (35.0)& 3.9 (30.0)\\
    ResNet50 (AT)$\xrightarrow{\text{KDIGA}}$ViT-S/16$^*$& 77.7 (63.1)& 65.3 (47.5)& 32.6 (42.5)& 13.4 (35.0)& 1.1 (30.0)\\
    \bottomrule[1pt]
    \end{tabular}
}
\end{table}

\vspace{-0.1cm}
\paragraph{Adversarial Robustness Can Transfer Between CNNs and Vision Transformers}
Vision transformers and CNNs have entirely different model architectures, while Table~\ref{tab:pgd-imagnet} and Table~\ref{tab:auto-imagnet} show that the adversarial robustness can be transferred between them with KDIGA. We have already shown that the adversarial robustness can be transferred from ViTs to CNNs in the previous paragraph, and here we show the reverse also holds. With PGD attack radius of $\{0.003, 0.005, 0.1\}$, the robust accuracy of ViT/S-16 obtained by ST is only $\{24.6\%, 10.2\%, 1.0\%\}$ while the distilled ViT-S/16 has an accuracy of $\{50.4\%, 33.5\%, 11.1\%\}$. While under AutoAttack, when the attack radius is $\{0.001, 0.003, 0.005\}$, the robust accuracy of ViT-S/16 (ST) is $\{14.3\%, 0.4\%, 0.0\%\}$, but after distillation, the robust accuracy becomes $\{65.3\%, 32.6\%, 13.4\%\}$. As ViTs are difficult to train even in the standard setting, this result shows that we can consider to transfer adversarial robustness from adversarially trained CNNs to ViTs to obtain robust ViTs.

\vspace{-0.1cm}
\paragraph{Input gradient alignment Works for Pre-trained Models}
The student model in standard knowledge distillation is generally trained from scratch. While in the experiments of ``ResNet50$\xrightarrow{KDIGA}$ViT-S/16$^\star$'' as shown in Table~\ref{tab:pgd-imagnet} and Table~\ref{tab:auto-imagnet}, we take the pre-trained ViT as the student to help the training converge in a shorter time. In Table~\ref{tab:auto-imagnet}, ViT-S/16 remains the high clean accuracy of $77.7\%$ after distillation compared with the original clean accuracy of $77.6\%$ in standard pre-training. This result shows the feasibility to further promote the adversarial robustness of a pre-trained model using our proposed input gradient alignment without harming the model's performance on the clean dataset in knowledge distillation, which gives a novel and inspiring approach to train new robust models more efficiently at less cost of the clean accuracy. 

\vspace{-0.1cm}
\paragraph{Students Can Obtain Even Better Adversarial Robustness Than Teachers}
From Proposition\ref{pro:robustness}, we proved that the student model can achieve at least the same certified robustness as the teacher model's under certain assumptions. The results in Table~\ref{tab:pgd-imagnet} and Table~\ref{tab:auto-imagnet} also show that the student can obtain even higher robust accuracy against adversarial perturbations than teacher's with KDIGA empirically. For example, when the teacher model is not robust, i.e. DEIT-S/16, ResNet18 distilled from it with KDIGA achieves an robust accuracy of $\{50.0\%, 23.7\%, 7.8\%\}$ against AutoAttack with attack radius of $\{0.001, 0.003, 0.005\}$, while the teacher model only has an accuracy of $\{0.4\%, 0.0\%. 0.0\%\}$ in the same situations. This shows that input gradient alignment can still help the student obtain better adversarial robustness even when the teacher is not very robust.

\vspace{-0.1cm}
\paragraph{Students with Higher Learning Capacity Achieve Better Results}
In Table~\ref{tab:pgd-imagnet}, we set the student model to different sizes, i.e., ResNet18, ResNet34, ResNet50, and ResNet101, to check the effect of the student's learning capacity on knowledge distillation. As shown in the table, both the clean accuracy and the robust accuracy increase as the the model size grows, e.g., ResNet101 achieves a robust accuracy of $12.7\%$ under attack radius of $0.01$ while ResNet18 achieves $9.2\%$ in the same case. Therefore, we can expect students with higher learning capacities to achieve better performance. It is also worth noting that the student model in KD is commonly smaller than the teacher for some practical purposes like model compression. While in our setting, student can be larger than the teacher model (like the case of noisy student training), meaning we can train a robust teacher with less computing cost and then transfer the adversarial robustness to larger students, in order to eventually obtain a high-capacity and robust model.

\subsection{Results on the CIFAR-10 Dataset}\label{sec:cifar}
We compare our method with ST, KD and ARD on the CIFAR-10 dataset in Table~\ref{tab:pgd-cifar}. We show that KDIGA has the highest clean accuracy compared with other baseline methods. We also show that two combinations of our method with ARD have the highest robust accuracy in knowledge distillation. 

We use the fair comparison setting described in Section~\ref{sec:setting}, though we find ARD can achieve higher robust accuracy with $\lambda_{CE}$ set to $0$ and $\lambda_{KL}$ set to $1$, in which case its clean accuracy will decrease a lot. $\lambda_{CE}$ is critical to preserve the clean accuracy and it is an important term in the standard KD. So we set both $\lambda_{CE}$ and $\lambda_{KL}$ to $0.5$ as in the standard KD setting for all experiments. 

From Table~\ref{tab:pgd-cifar}, we find the student distilled with KDIGA achieves comparable robust accuracy with ARD. But KDIGA has the advantage of dispensing the computation of the adversarial examples and thus is more cheap and efficient. Moreover, KDIGA has the highest clean accuracy which even exceeds the result of KD. When combined with KDIGA, both the KDIGA-ARD$_\text{C}$ and KDIGA-ARD$_\text{A}$ obtain higher robust accuracy than ARD. This shows that it is feasible to combine KDIGA with other state-of-the-art methods to further improve the student's robust accuracy.

\begin{table}[t]
    \centering
    \caption{Robust accuracy ($\%$) of student models against 20-step PGD attack with different radii and clean accuracy ($\%$) on the CIFAR-10 dataset. All students are trained from scratch. ``ST'' means the model is trained following the standard approach without distillation nor adversarial training. ``TRADES'' means the model is adversarially trained using TRADES~\citep{trades}.}
        \label{tab:pgd-cifar}
            \vspace{-2mm}
    \resizebox{\columnwidth}{!}{
    \begin{tabular}{l c c c c c c c c r}\toprule[1pt]
     & &\multicolumn{4}{c}{\bf{PGD Attack radius}}\\
    \bf{Model} &\bf{Clean} &\bf{1/255} &\bf{2/255} &\bf{3/255} &\bf{4/255} &\bf{5/255} &\bf{6/255} &\bf{7/255} &\bf{8/255}\\ \midrule
    WideResNet(TRADES)& 84.92 &82.36& 79.35& 75.99& 72.28& 68.54& 64.54& 60.57& 56.68 \\
    MobileNetV2 (ST)& 91.82& 7.43& 1.11& 0.04& 0.00& 0.00& 0.00& 0.00& 0.00\\
    WideResNet(TRADES)$\xrightarrow{\text{KD}}$MobileNetV2& 92.56& 42.74& 24.63& 16.79& 12.55& 9.98& 8.56& 7.17& 5.97\\
    WideResNet(TRADES)$\xrightarrow{\text{ARD}}$MobileNetV2& 91.65& 80.35& 68.34& 58.43& 49.49& 41.74& 33.78& 26.17& 20.73\\
    WideResNet(TRADES)$\xrightarrow{\text{KDIGA}}$MobileNetV2&$\bm{93.03}$ &60.68& 44.49& 36.29& 30.16& 25.68& 22.39& 19.76& 17.81\\
    WideResNet(TRADES)$\xrightarrow{\text{KDIGA-ARD}_\text{C}}$MobileNetV2& 92.22& 83.29& 71.76& 61.85& 53.46& 45.28& 37.70& 31.12& $\bm{25.85}$\\
    WideResNet(TRADES)$\xrightarrow{\text{KDIGA-ARD}_\text{A}}$MobileNetV2&  90.67& 81.82& 70.57& 60.69& 52.58& 45.50& 38.59& 32.72& $\bm{27.50}$\\
    \bottomrule[1pt]
    \end{tabular}
}

\end{table}

\subsection{Local Linearity Bounds for Adversarial Robustness in Knowledge Distillation}
From Proposition~\ref{pro:bound}, we prove that the certified robustness of the student model can be bounded by the LLM (as defined in Definition~\ref{def:llm}), the cross-entropy loss, and the gradient alignment norm, if we regard other terms of the teacher as fixed. Table~\ref{tab:llm} shows the bounds for adversarial robustness of models trained on CIFAR-10. 
We randomly sample 1000 test samples to calculate the terms in the bounds.

In reference to Table~\ref{tab:pgd-cifar} and Table~\ref{tab:llm}, the empirical performance matches the theoretical insights that models with better adversarial robustness have smaller values in the bounds, i.e., MobileNetV2 trained with standard training has the highest bounds, and students distilled with KDIGA-ARD has the lowest bounds. The LLM bound and input gradient alignment norm for ARD are much lower than KD, showing that adversarial training also has the effect of improving the local linearity and aligning the input gradients. KDIGA achieves similar bounds as ARD though the training process does not use adversarial examples. Table~\ref{tab:llm} also shows that combining our method with ARD can further reduce the bounds and induce better adversarial robustness. KD only has the lowest cross-entropy loss while other terms are high, which can explain its failure in preserving adversarial robustness, as its objective design only focuses on improving standard accuracy.

\begin{table}[t]
    \centering
\caption{Bounds for adversarial robustness (as defined in Proposition~\ref{pro:bound}) of different models on CIFAR-10. $llm_{\epsilon}$ is defined by Definition~\ref{def:llm} where $\epsilon$ is the radius of perturbations. $l_{CE}$ is the cross-entropy loss. $\|g^s-g^t\|_2$ calculates the $l_2$-norm of input gradient alignment term. ``ST'' means the model is trained following the standard approach without distillation nor adversarial training. ``TRADES'' means the model is adversarially trained using TRADES~\citep{trades}. We calculate $\|g^s-g^t\|_2$ with WideResNet(TRADES) as the teacher model for MobileNetV2 (ST) for comparison, though the training process of MobileNetV2 (ST) doesn't involve a teacher model.}
        \label{tab:llm}
          \vspace{-2mm}
    \resizebox{\columnwidth}{!}{
    \begin{tabular}{l c c c r}\toprule[1pt]
    Model &$llm_{4/255}$ & $llm_{8/255}$ & $l_{CE}$ & $\|g^s-g^t\|_2$\\ \midrule
    MobileNetV2 (ST)& 12.413 &21.691& 0.364& 4.099\\
    WideResNet(TRADES)$\xrightarrow{\text{KD}}$MobileNetV2& 5.960& 10.286& $\bm{0.218}$& 1.958\\
    WideResNet(TRADES)$\xrightarrow{\text{ARD}}$MobileNetV2& 1.326& 3.034& 0.261& 0.569\\
    WideResNet(TRADES)$\xrightarrow{\text{KDIGA}}$MobileNetV2& 2.561& 4.914& 0.235& 0.587\\
    WideResNet(TRADES)$\xrightarrow{\text{KDIGA-ARD}_\text{C}}$MobileNetV2& $\bm{1.081}$& $\bm{2.421}$& 0.228& $\bm{0.339}$\\
    WideResNet(TRADES)$\xrightarrow{\text{KDIGA-ARD}_\text{A}}$MobileNetV2& 1.107& 2.442& 0.285& 0.377\\
    \bottomrule[1pt]
    \end{tabular}
}

\end{table}

\section{Conclusion}
This paper provides a comprehensive study on how and when can adversarial robustness transfer from the teacher model to student model in knowledge distillation, in addition to standard accuracy. For the \textit{how}, we show that standard knowledge distillation fails to preserve adversarial robustness, and we propose a novel input gradient alignment technique (KDIGA) to address this issue.  For the \textit{when}, under certain assumptions we prove that using KDIGA the student model can be at least as robust as the teacher model, and we generalize our theoretical analysis using local linearity measures. The superior performance of KDIGA over baselines in terms of improved adversarial robustness while retaining clean accuracy is empirically validated using CNNs and vision transformers.

\bibliography{iclr2021_conference}

\begin{thebibliography}{33}
\providecommand{\natexlab}[1]{#1}
\providecommand{\url}[1]{\texttt{#1}}
\expandafter\ifx\csname urlstyle\endcsname\relax
  \providecommand{\doi}[1]{doi: #1}\else
  \providecommand{\doi}{doi: \begingroup \urlstyle{rm}\Url}\fi

\bibitem[Chan et~al.(2020)Chan, Tay, and Ong]{chan2020thinks}
Alvin Chan, Yi~Tay, and Yew-Soon Ong.
\newblock What it thinks is important is important: Robustness transfers
  through input gradients.
\newblock In \emph{Proceedings of the IEEE/CVF Conference on Computer Vision
  and Pattern Recognition}, pp.\  332--341, 2020.

\bibitem[Chen et~al.(2020)Chen, Liu, Chang, Cheng, Amini, and
  Wang]{chen2020adversarial}
Tianlong Chen, Sijia Liu, Shiyu Chang, Yu~Cheng, Lisa Amini, and Zhangyang
  Wang.
\newblock Adversarial robustness: From self-supervised pre-training to
  fine-tuning.
\newblock In \emph{Proceedings of the IEEE/CVF Conference on Computer Vision
  and Pattern Recognition}, pp.\  699--708, 2020.

\bibitem[Croce \& Hein(2020)Croce and Hein]{auto-attack}
Francesco Croce and Matthias Hein.
\newblock Reliable evaluation of adversarial robustness with an ensemble of
  diverse parameter-free attacks.
\newblock In \emph{International Conference on Machine Learning}, pp.\
  2206--2216. PMLR, 2020.

\bibitem[Croce et~al.(2019)Croce, Andriushchenko, and Hein]{linearity-max}
Francesco Croce, Maksym Andriushchenko, and Matthias Hein.
\newblock Provable robustness of relu networks via maximization of linear
  regions.
\newblock In \emph{the 22nd International Conference on Artificial Intelligence
  and Statistics}, pp.\  2057--2066. PMLR, 2019.

\bibitem[Deng et~al.(2009)Deng, Dong, Socher, Li, Li, and Fei-Fei]{imagenet}
Jia Deng, Wei Dong, Richard Socher, Li-Jia Li, Kai Li, and Li~Fei-Fei.
\newblock Imagenet: A large-scale hierarchical image database.
\newblock In \emph{2009 IEEE conference on computer vision and pattern
  recognition}, pp.\  248--255. Ieee, 2009.

\bibitem[Dosovitskiy et~al.(2020)Dosovitskiy, Beyer, Kolesnikov, Weissenborn,
  Zhai, Unterthiner, Dehghani, Minderer, Heigold, Gelly, et~al.]{vit}
Alexey Dosovitskiy, Lucas Beyer, Alexander Kolesnikov, Dirk Weissenborn,
  Xiaohua Zhai, Thomas Unterthiner, Mostafa Dehghani, Matthias Minderer, Georg
  Heigold, Sylvain Gelly, et~al.
\newblock An image is worth 16x16 words: Transformers for image recognition at
  scale.
\newblock \emph{arXiv preprint arXiv:2010.11929}, 2020.

\bibitem[Engstrom et~al.(2019)Engstrom, Ilyas, Salman, Santurkar, and
  Tsipras]{imagenet-robsut-cnn}
Logan Engstrom, Andrew Ilyas, Hadi Salman, Shibani Santurkar, and Dimitris
  Tsipras.
\newblock Robustness (python library), 2019.
\newblock URL \url{https://github.com/MadryLab/robustness}.

\bibitem[Goldblum et~al.(2020)Goldblum, Fowl, Feizi, and Goldstein]{ard}
Micah Goldblum, Liam Fowl, Soheil Feizi, and Tom Goldstein.
\newblock Adversarially robust distillation.
\newblock In \emph{Proceedings of the AAAI Conference on Artificial
  Intelligence}, volume~34, pp.\  3996--4003, 2020.

\bibitem[Gou et~al.(2021)Gou, Yu, Maybank, and Tao]{distillation1}
Jianping Gou, Baosheng Yu, Stephen~J Maybank, and Dacheng Tao.
\newblock Knowledge distillation: A survey.
\newblock \emph{International Journal of Computer Vision}, 129\penalty0
  (6):\penalty0 1789--1819, 2021.

\bibitem[He et~al.(2016)He, Zhang, Ren, and Sun]{resnet}
Kaiming He, Xiangyu Zhang, Shaoqing Ren, and Jian Sun.
\newblock Deep residual learning for image recognition.
\newblock In \emph{Proceedings of the IEEE conference on computer vision and
  pattern recognition}, pp.\  770--778, 2016.

\bibitem[Hendrycks et~al.(2019)Hendrycks, Lee, and Mazeika]{hendrycks2019using}
Dan Hendrycks, Kimin Lee, and Mantas Mazeika.
\newblock Using pre-training can improve model robustness and uncertainty.
\newblock In \emph{International Conference on Machine Learning}, pp.\
  2712--2721. PMLR, 2019.

\bibitem[Hinton et~al.(2015)Hinton, Vinyals, and Dean]{hinton2015distilling}
Geoffrey Hinton, Oriol Vinyals, and Jeff Dean.
\newblock Distilling the knowledge in a neural network.
\newblock \emph{arXiv preprint arXiv:1503.02531}, 2015.

\bibitem[Krizhevsky et~al.()Krizhevsky, Nair, and Hinton]{cifar10}
Alex Krizhevsky, Vinod Nair, and Geoffrey Hinton.
\newblock Cifar-10 (canadian institute for advanced research).
\newblock URL \url{http://www.cs.toronto.edu/~kriz/cifar.html}.

\bibitem[Lee et~al.(2019)Lee, Alvarez-Melis, and Jaakkola]{linearity-towards}
Guang-He Lee, David Alvarez-Melis, and Tommi~S Jaakkola.
\newblock Towards robust, locally linear deep networks.
\newblock \emph{arXiv preprint arXiv:1907.03207}, 2019.

\bibitem[Lyu \& Chen(2020)Lyu and Chen]{lyu2020differentially}
Lingjuan Lyu and Chi-Hua Chen.
\newblock Differentially private knowledge distillation for mobile analytics.
\newblock In \emph{Proceedings of the 43rd International ACM SIGIR Conference
  on Research and Development in Information Retrieval}, pp.\  1809--1812,
  2020.

\bibitem[Madry et~al.(2017)Madry, Makelov, Schmidt, Tsipras, and
  Vladu]{pgd-attack}
Aleksander Madry, Aleksandar Makelov, Ludwig Schmidt, Dimitris Tsipras, and
  Adrian Vladu.
\newblock Towards deep learning models resistant to adversarial attacks.
\newblock \emph{arXiv preprint arXiv:1706.06083}, 2017.

\bibitem[Mirzadeh et~al.(2020)Mirzadeh, Farajtabar, Li, Levine, Matsukawa, and
  Ghasemzadeh]{mirzadeh2020improved}
Seyed~Iman Mirzadeh, Mehrdad Farajtabar, Ang Li, Nir Levine, Akihiro Matsukawa,
  and Hassan Ghasemzadeh.
\newblock Improved knowledge distillation via teacher assistant.
\newblock In \emph{Proceedings of the AAAI Conference on Artificial
  Intelligence}, volume~34, pp.\  5191--5198, 2020.

\bibitem[Naseer et~al.(2021)Naseer, Ranasinghe, Khan, Hayat, Khan, and
  Yang]{robust_vit3}
Muzammal Naseer, Kanchana Ranasinghe, Salman Khan, Munawar Hayat, Fahad~Shahbaz
  Khan, and Ming-Hsuan Yang.
\newblock Intriguing properties of vision transformers.
\newblock \emph{arXiv preprint arXiv:2105.10497}, 2021.

\bibitem[Passban et~al.(2020)Passban, Wu, Rezagholizadeh, and
  Liu]{passban2020alp}
Peyman Passban, Yimeng Wu, Mehdi Rezagholizadeh, and Qun Liu.
\newblock Alp-kd: Attention-based layer projection for knowledge distillation.
\newblock \emph{arXiv preprint arXiv:2012.14022}, 2020.

\bibitem[Paul \& Chen(2021)Paul and Chen]{robust_vit2}
Sayak Paul and Pin-Yu Chen.
\newblock Vision transformers are robust learners.
\newblock \emph{arXiv preprint arXiv:2105.07581}, 2021.

\bibitem[Qin et~al.(2019)Qin, Martens, Gowal, Krishnan, Dvijotham, Fawzi, De,
  Stanforth, and Kohli]{llm}
Chongli Qin, James Martens, Sven Gowal, Dilip Krishnan, Krishnamurthy
  Dvijotham, Alhussein Fawzi, Soham De, Robert Stanforth, and Pushmeet Kohli.
\newblock Adversarial robustness through local linearization.
\newblock \emph{arXiv preprint arXiv:1907.02610}, 2019.

\bibitem[Sandler et~al.(2018)Sandler, Howard, Zhu, Zhmoginov, and
  Chen]{mobilenetv2}
Mark Sandler, Andrew Howard, Menglong Zhu, Andrey Zhmoginov, and Liang-Chieh
  Chen.
\newblock Mobilenetv2: Inverted residuals and linear bottlenecks.
\newblock In \emph{Proceedings of the IEEE conference on computer vision and
  pattern recognition}, pp.\  4510--4520, 2018.

\bibitem[Sattelberg et~al.(2020)Sattelberg, Cavalieri, Kirby, Peterson, and
  Beveridge]{linearity-local}
Ben Sattelberg, Renzo Cavalieri, Michael Kirby, Chris Peterson, and Ross
  Beveridge.
\newblock Locally linear attributes of relu neural networks.
\newblock \emph{arXiv preprint arXiv:2012.01940}, 2020.

\bibitem[Shafahi et~al.(2019)Shafahi, Saadatpanah, Zhu, Ghiasi, Studer, Jacobs,
  and Goldstein]{shafahi2019adversarially}
Ali Shafahi, Parsa Saadatpanah, Chen Zhu, Amin Ghiasi, Christoph Studer, David
  Jacobs, and Tom Goldstein.
\newblock Adversarially robust transfer learning.
\newblock \emph{arXiv preprint arXiv:1905.08232}, 2019.

\bibitem[Shao et~al.(2021)Shao, Shi, Yi, Chen, and Hsieh]{robust_vit1}
Rulin Shao, Zhouxing Shi, Jinfeng Yi, Pin-Yu Chen, and Cho-Jui Hsieh.
\newblock On the adversarial robustness of visual transformers.
\newblock \emph{arXiv preprint arXiv:2103.15670}, 2021.

\bibitem[Sun et~al.(2019)Sun, Cheng, Gan, and Liu]{sun2019patient}
Siqi Sun, Yu~Cheng, Zhe Gan, and Jingjing Liu.
\newblock Patient knowledge distillation for bert model compression.
\newblock \emph{arXiv preprint arXiv:1908.09355}, 2019.

\bibitem[Wang et~al.(2019)Wang, Bao, Sun, Zhu, Cao, and
  Philip]{wang2019private}
Ji~Wang, Weidong Bao, Lichao Sun, Xiaomin Zhu, Bokai Cao, and S~Yu Philip.
\newblock Private model compression via knowledge distillation.
\newblock In \emph{Proceedings of the AAAI Conference on Artificial
  Intelligence}, volume~33, pp.\  1190--1197, 2019.

\bibitem[Wang et~al.(2021{\natexlab{a}})Wang, Xu, Liu, Chen, Weng, Gan, and
  Wang]{wang2021on}
Ren Wang, Kaidi Xu, Sijia Liu, Pin-Yu Chen, Tsui-Wei Weng, Chuang Gan, and Meng
  Wang.
\newblock On fast adversarial robustness adaptation in model-agnostic
  meta-learning.
\newblock In \emph{International Conference on Learning Representations},
  2021{\natexlab{a}}.
\newblock URL \url{https://openreview.net/forum?id=o81ZyBCojoA}.

\bibitem[Wang et~al.(2021{\natexlab{b}})Wang, Li, Shi, Xian, and
  Cao]{wang2021knowledge}
Yiran Wang, Xingyi Li, Min Shi, Ke~Xian, and Zhiguo Cao.
\newblock Knowledge distillation for fast and accurate monocular depth
  estimation on mobile devices.
\newblock In \emph{Proceedings of the IEEE/CVF Conference on Computer Vision
  and Pattern Recognition}, pp.\  2457--2465, 2021{\natexlab{b}}.

\bibitem[Wightman(2019)]{timm}
Ross Wightman.
\newblock Pytorch image models.
\newblock \url{https://github.com/rwightman/pytorch-image-models}, 2019.

\bibitem[Zagoruyko \& Komodakis(2016{\natexlab{a}})Zagoruyko and
  Komodakis]{wideresnet}
Sergey Zagoruyko and Nikos Komodakis.
\newblock Wide residual networks.
\newblock \emph{arXiv preprint arXiv:1605.07146}, 2016{\natexlab{a}}.

\bibitem[Zagoruyko \& Komodakis(2016{\natexlab{b}})Zagoruyko and
  Komodakis]{zagoruyko2016paying}
Sergey Zagoruyko and Nikos Komodakis.
\newblock Paying more attention to attention: Improving the performance of
  convolutional neural networks via attention transfer.
\newblock \emph{arXiv preprint arXiv:1612.03928}, 2016{\natexlab{b}}.

\bibitem[Zhang et~al.(2019)Zhang, Yu, Jiao, Xing, El~Ghaoui, and
  Jordan]{trades}
Hongyang Zhang, Yaodong Yu, Jiantao Jiao, Eric Xing, Laurent El~Ghaoui, and
  Michael Jordan.
\newblock Theoretically principled trade-off between robustness and accuracy.
\newblock In \emph{International Conference on Machine Learning}, pp.\
  7472--7482. PMLR, 2019.

\end{thebibliography}
\bibliographystyle{iclr2021_conference}

\clearpage
\appendix
\section*{Appendix}

\section{Proof for Proposition~\ref{pro:robustness}}\label{app:proof1}

Since $f^t$ is $\delta$-robust, the prediction of $f^t(\vx)$ is invariant to the input perturbations smaller than the certified robust radius by definition, i.e., 

\begin{equation}
    \argmax\ f^t(\vx + \bm{\epsilon}) = \argmax\ f^t(\vx), \ \ \forall \vx \in \mathcal{D}, \ \ \forall \bm{\epsilon} \in (0, \delta)^D,
\end{equation}
where $\mathcal{D}$ is the task-specific data set. 
Denote the student model distilled from the teacher model using normal knowledge distillation as $f^{KD}(\vx): \R^D \rightarrow \R^N$. The loss of the normal knowledge distillation can be formulated as
\begin{equation}
    \gL_{KD}(\vx, y) = \lambda_{CE}\gL_{CE}(f^{KD}(\vx), y) + \lambda_{KL} T^2 \gL_{KL}(f^{KD}(\vx)/T, f^t(\vx)/T), \ \ \forall (\vx, y) \in \mathbb{D}, \label{eq:loss-kd}
\end{equation}
where $\gL_{CE}$ is the cross-entropy loss, $\gL_{KL}$ is the KL-divergence loss which is also called the soft loss in knowledge distillation, $T$ is the temperature factor, and $\lambda_{CE}, \lambda_{KL}$ are hyper-parameters to balance the effects of the two losses.
The loss of KDIGA is calculated by
\begin{equation}
\begin{aligned}
    \gL_{IGA}(\vx, y) =&\lambda_{CE}\gL_{CE}(f^{IGA}(\vx), y) + \lambda_{KL} T^2 \gL_{KL}(f^{IGA}(\vx)/T, f^t(\vx)/T)\\
    &+ \lambda_{IGA} \|\nabla_{\vx} \gL_{CE}(f^{IGA}(\vx), y) - \nabla_{\vx}\gL_{CE}(f^t(\vx), y)\|_2, \ \ \forall (\vx, y) \in \mathbb{D}, \label{eq:loss-iga}
\end{aligned}
\end{equation}
where $f^{IGA}$ is the student model, $\lambda_{CE}, \lambda_{KL}$ and $\lambda_{IGA}$ are hyper-parameters.

Without loss of generality, we set the temperature factor $T=1$ for both KD and KDIGA. According to the perfect student assumption, $f^{IGA}$ satisfies the following equations: 
\begin{numcases}{}
    \nabla_{\vx} \gL_{IGA} (\vx, y) - \nabla_{\vx} \gL_{IGA}(\vx, y) = 0 & \label{eq:gradient} \\
    f^{IGA}(\vx) - f^t(\vx) = 0 & \label{eq:assum-logit} \\
    f^{IGA}(\vx) = y, & $\forall (\vx, y) \in \mathcal{D}.$
\end{numcases}
The cross-entropy loss is defined as
\begin{equation}
\begin{aligned}
    \gL_{CE}(f(\vx), y) = -\log \big (\frac{\exp(f(\vx)_y)}{\sum_j \exp(f(\vx)_j)} \big ) = -f(\vx)_y + \log(\sum_j \exp(f(\vx)_j)),
\end{aligned}
\end{equation}
where $f(\cdot)$ is a classifier and $f(\vx)_j$ is the $j$-th prediction of the output.
Then the gradient of the cross-entropy loss with respect to the input is

\begin{equation}
    \begin{aligned}
        \nabla_{\vx} \gL_{CE}(f(\vx), y) &= -\nabla_{\vx} f(\vx)_y + \nabla_{\vx} \log (\sum_j \exp(f(\vx)_j)) \\
        &= -\nabla_{\vx} f(\vx)_y + \frac{\nabla_{\vx}(\sum_i \exp(f(\vx)_i))}{\sum_j \exp(f(\vx)_j)} \\
        &= -\nabla_{\vx} f(\vx)_y + \frac{\sum_i \nabla_{\vx} \exp(f(\vx)_i)}{\sum_j \exp(f(\vx)_j)}\\
        &= -\nabla_{\vx} f(\vx)_y + \frac{\sum_i \exp(f(\vx)_i) \nabla_{\vx}f(\vx)_i}{\sum_j \exp(f(\vx)_j)} 
    \end{aligned}
\end{equation}

Denote $\vg=g(\vx) = \nabla_{\vx}f(\vx)$, $\bm{\alpha} = \text{softmax} (f(\vx))$, then

\begin{equation}
    \begin{aligned}
        \nabla_{\vx} \gL_{CE}(f(\vx), y) &= -g(\vx)_y + \frac{\sum_i \exp(f(\vx)_i) g(\vx)_i}{\sum_j \exp(f_j(\vx))} \\
        &= -g(\vx)_y + \bm{\alpha} \cdot \vg \\
        &= (\bm{\alpha} - \vi_y)\cdot \vg.
    \end{aligned}
\end{equation}
where $\vi_y = (0, \cdots, 0, 1, 0, \cdots, 0)$ is an unit vector of which the $y$-th element equals one.
According to Eq.~\ref{eq:assum-logit}, $\bm{\alpha}^t = \bm{\alpha}^{IGA} = \bm{\alpha}$. The third term in Eq.~\ref{eq:loss-iga} for input gradient alignment is
\begin{equation}
    \begin{aligned}
        &\| \nabla_{\vx} \gL_{CE}(f^{t}(\vx), y) - \nabla_{\vx} \gL_{CE}(f^{IGA}(\vx), y) \| \\ =& \| (\bm{\alpha}^t - \vi_y) \cdot \vg^t - (\bm{\alpha}^{IGA} - \vi_y) \cdot \vg^{IGA} \| \\
        =& \| (\bm{\alpha} - \vi_y) \cdot (\vg^t - \vg^{IGA})\|.
    \end{aligned}
\end{equation}

Given $\bm{\alpha} - \vi_y \neq \bm{0}$, $\vg^t - \vg^{IGA}$ must be $\bm{0}$ since $\bm{\alpha} - \vi_y$ and $\vg^t - \vg^{IGA}$ are not strictly orthogonal unless $\vg^t - \vg^{IGA} = \bm{0}$. According to Eq.~\ref{eq:gradient},  we have $\vg^t - \vg^{IGA} = \bm{0}$.

According to the local linearity assumption,$\forall \vx \in \mathbb{D}$, $\forall \bm{\epsilon} \in [0, \delta)^{H \times W \times C}$,
\begin{equation}
    \begin{aligned}
        f^{IGA}(\vx + \bm{\epsilon}) &= f^{IGA}(\vx) + \bm{\epsilon}^T \cdot g^{IGA}(\vx) \\
        &= f^t(\vx) + \bm{\epsilon}^T \cdot g^t(\vx) \\
        &= f^t(\vx + \bm{\epsilon}) = f^t(\vx) = f^{IGA}(\vx).
    \end{aligned}
\end{equation}

Therefore, the certified robust radius of $f^{IGA}$ is at least $\delta$, which proves Proposition~\ref{pro:robustness}.

However, the knowledge distillation without input gradient alignment cannot guarantee the adversarial robustness preservation. Suppose $f^{KD}$ is a perfect student, we have
\begin{numcases}{}
    f^{KD}(\vx) - f^t(\vx) = 0 & \label{eq:kd-logit}\\
    f^{KD}(\vx) = y, & $\forall (\vx, y) \in \mathcal{D}.$
\end{numcases}
We point out that $f^{KD}$ can have different predictions around $\vx$, for example, let $\Tilde{\vx} = \vx + \bm{\epsilon} \in \mathcal{\text{\r{B}}} (\vx, \delta)$, denote $h(\vx) = f^{KD}(\vx) - f^t(\vx)$, then $h(\vx) = 0$, $\forall (\vx, y) \in \mathcal{D}$ according to Eq.~\ref{eq:kd-logit}. But $\exists h(\vx)$, $\exists \vx \in \mathcal{\text{\r{B}}} (\vx, \delta)$ s.t.
\begin{equation}
    \argmax f^{KD}(\vx) \neq \argmax f^t(\vx)
\end{equation}
since the first-order derivative of $h(\vx)$ is not constrained to be $0$ in the neighbourhood of $\vx$. This means the predictions of the student model distilled using knowledge distillation without input gradient alignment can be altered if we add perturbations to the input image.

\section{Proof for Proposition~\ref{pro:bound}}\label{app:proof2}

\begin{equation}
\begin{aligned}
&\left| \mathcal{L}_{CE} (f^s(\vx+\bm{\epsilon}),y) - \mathcal{L}_{CE}(f^t(\vx+\bm{\epsilon}),y) \right| \\
=& \left| \mathcal{L}_{CE}(f^s(\vx+\bm{\epsilon}),y) - \mathcal{L}_{CE}(f^s(\vx),y)-\bm{\epsilon}^T\nabla_{\vx}\mathcal{L}_{CE}(f^s(\vx),y) \right.\\
& - \left( \mathcal{L}_{CE}(f^t(\vx+\bm{\epsilon}),y) - \mathcal{L}_{CE}(f^t(\vx),y)-\bm{\epsilon}^T\nabla_{\vx}\mathcal{L}_{CE}(f^t(\vx),y) \right) \\
& +\left( \mathcal{L}_{CE}(f^s(\vx),y)-\mathcal{L}_{CE}(f^t(\vx),y) \right)\\
& + \bm{\epsilon}^T\left( \nabla_{\vx}\mathcal{L}_{CE}(f^s(\vx),y) - \nabla_{\vx}\mathcal{L}_{CE}(f^t(\vx),y) \right)\left.\right|\\
\leq & \max_{\bm{\epsilon}\in B(\delta)}
\left| \mathcal{L}_{CE}(f^s(\vx+\bm{\epsilon}),y) - \mathcal{L}_{CE}(f^s(\vx),y)-\bm{\epsilon}^T\nabla_{\vx}\mathcal{L}_{CE}(f^s(\vx),y) \right|\\
&+ \max_{\bm{\epsilon}\in B(\delta)}
\left| \mathcal{L}_{CE}(f^t(\vx+\bm{\epsilon}),y) - \mathcal{L}_{CE}(f^t(\vx),y)-\bm{\epsilon}^T\nabla_{\vx}\mathcal{L}_{CE}(f^t(\vx),y) \right| \\
&+ \mathcal{L}_{CE}(f^s(\vx),y) + \mathcal{L}_{CE}(f^t(\vx),y)
+ \delta \|\nabla_{\vx}\mathcal{L}_{CE}(f^s(\vx),y) - \nabla_{\vx}\mathcal{L}_{CE}(f^t(\vx),y)\|.
\end{aligned}
\end{equation}

\end{document}